\documentclass[10pt,twocolumn,letterpaper]{article}

\usepackage{cvpr}
\usepackage{times}
\usepackage{epsfig}
\usepackage{graphicx}
\usepackage{amsmath}
\usepackage{amssymb}
\usepackage{multirow}
\usepackage{adjustbox}

\usepackage[breaklinks=true,bookmarks=false]{hyperref}

\cvprfinalcopy 

\def\cvprPaperID{****} 

\begin{document}

\title{Fashion-IQ 2020 Challenge 2nd Place Team's Solution}

\author{Minchul Shin \\
Search Solutions\\
{\tt\small min.stellastra@gmail.com}
\and
Yoonjae Cho\\
NAVER/LINE Vision\\
{\tt\small yoonjae.cho@navercorp.com}
\and
Seongwuk Hong\\
NAVER/LINE Vision\\
{\tt\small seongwuk.hong@navercorp.com}
}

\maketitle

\begin{abstract}
This paper is dedicated to team VAA’s approach submitted to the Fashion-IQ challenge in CVPR 2020. Given a pair of the image and the text, we present a novel multimodal composition method, RTIC, that can effectively combine the text and the image modalities into a semantic space. We extract the image and the text features that are encoded by the CNNs and the sequential models (e.g., LSTM or GRU), respectively. To emphasize the meaning of the residual of the feature between the target and candidate, the RTIC is composed of N-blocks with channel-wise attention modules. Then, we add the encoded residual to the feature of the candidate image to obtain a synthesized feature. We also explored an ensemble strategy with variants of models and achieved a significant boost in performance comparing to the best single model. Finally, our approach achieved 2nd place in the Fashion-IQ 2020 Challenge with a test score of 48.02 on the leaderboard.
\end{abstract}

\section{Introduction}
The goal of the Fashion-IQ Challenge is to build an interactive image retrieval system from the given input image and text in the fashion domain. To be more specific, the input source image is drawn to the target image by the composition with the captions in text. The topic mentioned above suggests that user feedback can be applied to more intuitive searches and dramatically contributes to the new direction of the search system.

We propose Residual Text Image Composer (RTIC) inspired by recent works on image-text composition method TIRG~\cite{vo2019composing}: emphasizing the encoding of the residual between the candidate and the target features that can be modularized with multiple blocks. Each block encodes the residual for the specific channels of the feature using a channel-wise attention score for a gating mechanism. Our experimental results support that the proposed method, RTIC, can marginally outperform the existing baseline, TIRG, using the same hyper-parameter setting for training. Moreover, we propose an iterative process of the ensemble that uses the previous best score matrix as the new candidate of the ensemble. We used \textit{hyperopt}\footnote{https://github.com/hyperopt/hyperopt} Bayesian optimization~\cite{snoek2012practical} for the task, which aims to find optimal weights between similarity scores of the single models and maximize the score on the validation and the test. Finally, our approach achieves the 2nd place on the Fashion-IQ 2020 challenge with the ensemble of the discriminative models.

\begin{figure*}[t!]
\centering
\includegraphics[width=1.0\textwidth]{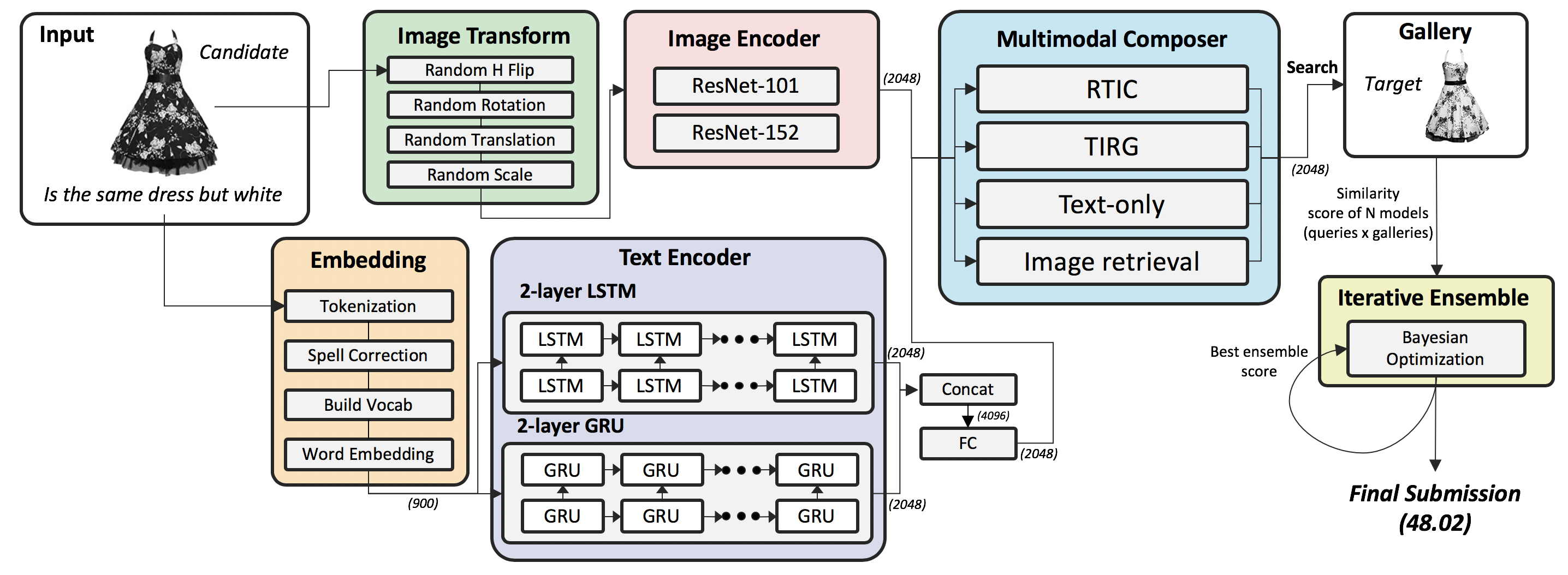}
\caption{The entire pipeline of our approach}
\label{fig:main_diagram}
\end{figure*}

\section{Method}
We describe encoders for text and image, followed by the single models used in the ensemble.
While composing the desired feature from different modalities is a crucial factor for the challenge, the method for encoding the text and the image significantly affects the final performance as well. Inspired by the winning team's solution of the last year~\cite{lastyear1st}, we used four variants of single models (\textit{Text-only, TIRG, RTIC, and IR-match}) with different characteristics to maximize the effectiveness of the ensemble. For clarity, we denote the nonlinear function of image encoder, text encoder, and multimodal composer as $\psi$, $\phi$, and $\lambda$ respectively. The image embedding is represented as $f^c_I = \psi(x^c_{image})$ and $f^t_I = \psi(x^t_{image})$ where $x^c_{image}$ and $x^t_{image}$ are image input of candidate and target. The sentence embedding is represented as $f_T = \phi(x_{text})$, where $x_{text}$ is text input. The composed embedding conditioned on $x_{text}$ is represented as $\Tilde{f} = \lambda(f^c_I; f_T)$. Our ultimate goal is to locate $\Tilde{f}$ in close distance with $f^t_{image}$ as possible.

\subsection{Text and image encoders}
We encoded the sentence embeddings from scratch using both LSTM~\cite{hochreiter1997long} and GRU~\cite{cho2014learning}. To be specific, the captions were firstly tokenized, and word embeddings were obtained using a bag of words. The word embeddings were initialized with GloVe~\cite{pennington2014glove} unless the tokenized word does not exist in GloVe vocabulary. The word embeddings were forwarded by 2-layered LSTM and GRU both, then the concatenation of the final hidden state output from LSTM and GRU was used for sentence embedding. We observe no improvement by using the average of word embeddings by GloVe for sentence embedding. For image encoder, we used ResNet-101 and ResNet-152. Both text and image encoders were trained from scratch without pretraining on any external data. For training the LSTM and the GRU from scratch, we initialize the word embedding with the concatenation of three GloVe vectors\footnote{https://github.com/stanfordnlp/GloVe} learned from \textit{Wikipedia}, \textit{Twitter}, and \textit{Common Crawl} that results in 900-dimensional input for the text encoder.

\subsection{Multimodal composer}

\noindent
\textbf{Text-only.}
Zhao and Ramanishka \textit{et al.}~\cite{lastyear1st} trained a model that ignores the candidate images based on the observation that 49\% of relative captions directly describe the target images. Inspired by its simplicity, we followed the same strategy. $\Tilde{f}$ is directly generated from $f_T$ using a simple encoder-decoder mechanism. Then triplet loss was used by constructing hard example pairs on the fly in mini batch.

\noindent
\textbf{Image retrieval.}
The instance retrieval (IR) model~\cite{noh2017large, gordo2016deep, gordo2017end} encodes the information of the visual similarity between objects to the feature. The IR representation powerful in that it enables the search engine to find the visually identical items among the million-scale images. Inspired by this, we trained IR model using DeepFashion~\cite{liu2016deepfashion} dataset and extracted the IR features off-the-shelf. Then we force the output of the image encoder and multimodal composers (e.g., TIRG or RTIC) to be the same with the extracted IR feature of a given candidate or target image. It is a knowledge transfer from the IR model to our image encoder and the multimodal composer instead of learning from scratch using pair-wise ranking loss.

\noindent
\textbf{TIRG.}
The TIRG~\cite{vo2019composing} is a baseline method that uses the image and the text information both to generate the desired feature. It is effective for the queries where the text describes modifications to the candidate images. We used TIRG and its variants as a baseline.

\noindent
\textbf{RTIC.}
The residual text and image composer (RTIC) is our proposed method that learns the residual between the features of target and candidate images. Given the features of candidate and target images extracted from image encoder, we consider the target $f^{t}_{I}$ as an addition of the candidate $f^{c}_{I}$ and the residual $h$ conditioned on the text $f_{T}$ that is obtained by $h = \rho(f^{c}_{I}; f_{T})$. Therefore the final composed embedding is formulated as $\Tilde{f} = \lambda(f^c_I; f_T) \simeq f^{c}_{I} + h= f^{t}_{I}$. The RTIC is a composition of $N$ blocks with channel-wise attention $A \in R^{d \times N}$, where d is the dimension of the image feature, and $N$ is the number of blocks. Every block is in charge of encoding the residual of the specific channels. The input $x$ of each block is computed with the dot product with the feature $f \in R^{d}$ outputted from the previous block and the attention score $A_i \in R^{d}$, where $A_i$ is the $i$-th channel score of $A$. The residual is ignored if the attention score is 0 for the specific channel functioning as a gate mechanism.

\begin{table*}[t!]
\caption{The model performance on the validation of the single and the ensemble. TIRG with ${\dag}$ mark indicates the model is trained under default setting with no ablation. Ensemble 4 models is an ensemble of Text-only, Image retrieval, TIRG, and RTIC.}
\centering
\vspace{2mm}
\begin{adjustbox}{width=1\textwidth}
\begin{tabular}{cccccccccc}
\hline
\multirow{2}{*}{Method} & \multirow{2}{*}{Text Encoder} & \multirow{2}{*}{Image Encoder} & \multirow{2}{*}{Average} & \multicolumn{2}{c}{Shirt} & \multicolumn{2}{c}{Dress} & \multicolumn{2}{c}{Toptee} \\
\cline{5-10}
 &  &  &  & R10 & R50 & R10 & R50 & R10 & R50 \\
\hline \hline
Text-only & LSTM+GRU & ResNet-152 & 23.87 & 11.53 & 31.21 & 12.30 & 32.42 & 15.55 & 40.18 \\
Image retrieval & LSTM+GRU & ResNet-101 & 33.68 & 19.23 & 41.12 & 22.41 & 43.08 & 24.83 & 51.40 \\
${\dag}$TIRG & - & - & 28.20 & 14.13 & 34.15 & 18.14 & 42.53 & 18.56 & 41.71 \\
TIRG & LSTM+GRU & ResNet-152 & 37.18 & 20.02 & 44.55 & 25.73 & 49.88 & 26.72 & 54.82 \\
RTIC & SWEM~\cite{shen2018baseline} & ResNet-101 & 34.06 & 19.09 & 41.32 & 23.55 & 46.90 & 24.68 & 48.85 \\
RTIC & LSTM & ResNet-101 & 37.22 & \textbf{21.39} & 44.50 & 26.47 & 50.47 & 26.98 & 53.49 \\
RTIC & LSTM+GRU & ResNet-101 & \textbf{38.22} & 21.30 & \textbf{44.80} & \textbf{28.21} & \textbf{51.41} & \textbf{28.00} & \textbf{55.58} \\
\hline
RTIC+Text-only & - & - & 40.12 & 22.77 & 34.74 & 28.36 & 53.64 & 30.29 & 58.95 \\
RTIC+Image retrieval & - & - & 42.66 & 25.17  & 49.02 & 30.44 & 56.92 & 33.96 & 60.43 \\
RTIC+TIRG & - & - & 40.78 & 23.06 & 47.69 & 29.60 & 54.73 & 30.70 & 58.95 \\
\hline
Ensemble 4 models & - & - & 45.05 & 26.55 & 52.65 & 33.07 & 59.35 & 35.49 & 63.23 \\
Ensemble N models iteratively & - & - & \textbf{47.55} & \textbf{29.69} & \textbf{54.71} & \textbf{35.01} & \textbf{62.02} & \textbf{37.99} & \textbf{65.88} \\
\hline
\end{tabular}
\end{adjustbox}
\label{tab:single_model_performance}
\end{table*}

\subsection{Spell Correction}
The Fashion-IQ dataset includes a lot of misspelled words (e.g., whtie $\rightarrow$ white). To fix the misspelled word, first, we tokenize the sentence to get a set of tokens and check if the token is found in the existing GloVe vocabulary. If the token is not found, we correct the spell using the \textit{pyspellchecker}\footnote{https://github.com/barrust/pyspellchecker} python package. Then we manually validated the corrected word is likely to be used in the fashion domain.

\subsection{Ensemble Strategy}
We obtain the similarity scores matrices $H_{s_i} \in R^{q \times g}$ between the query and gallery for every single model as the results, where $s_i$, $q$, and $g$ are the $i$-th single model, the number of queries and the galleries respectively. Given $H \in \{H_{s_1}, H_{s_2},... , H_{s_n}\}$, we find the set of $W \in \{w_1, w_2, ..., w_n\}$ to obtain the weighted sum of the scores $H$. The Bayesian optimization provided by the \textit{hyperopt} package is used for the task. First, $n$ uniform distribution (0, 1) was set as the weight search space for the ensemble given outputs of $n$ single models. The randomly sampled weights were fed to Tree-structured Parzen Estimator (TPE) algorithm~\cite{bergstra2011algorithms} that finds the optimal $W$ with the objective set to $f(W) = (R@10 + R@50)/2$. Finally, the output score in the ensemble is also a form of a score matrix $H_{best} = w_1H_1 + w_2H_2 + ... w_nH_n \in R^{q \times g}$. We treat the best score $H_{best}$ at the moment as an output of a single model and maximize the best score by including $H_{best}$ as a candidate for every trial of the ensemble. It is an iterative accumulation of the single model result. However, we found that the excessive accumulation breaks the tendency between the results on the test and the validation because the score gap between test and validation would be accumulated as well.

\subsection{Technical Details}
We used the triplet loss with hard example mining in the mini-batch~\cite{hermans2017defense}. We also investigated various ranking losses such as SoftTriple loss~\cite{qian2019softtriple}, Multi-Similarity loss~\cite{wang2019multi}, and Circle loss~\cite{sun2020circle}, but found no gain in performance. Note that we did not optimize the hyperparameters for each loss which means additional parameter optimization might provide better performance for the task. For the image transformation, the image was resized to 224 x 224 and padded maintaining the ratio followed by a random horizontal flip and a random affine including rotation, translation, and scaling. We observed no improvements by adding random erasing. We used the AdamW optimizer~\cite{loshchilov2017decoupled} with the default setting provided by PyTorch if not stated otherwise. We set the hyper-parameters $lr$ = 0.00011148, $beta$ = 0.47, and $batch size$ = 32, where $lr$ is the initial learning rate, $beta$ is a coefficient for AdamW. The learning rate is decayed with a factor of 0.474 every 10 epochs during the 80 epochs of training in total. We applied small $lr$ for the image encoder by lowering the $lr$ with a factor of 0.48. The image and the text encoder were not pretrained on any external datasets. To describe the preprocessing the texts, we add special token "[CLS]" at the beginning and join the sentences with special token "[SEP]". If the tokenized word is not found in our vocabulary, the word is set to the token "[UNK]". We randomly shuffle the order of the sentences every epoch for better generalization.

\begin{table}[t!]
\caption{Performance improvements of single model improvements on each training details. We changed optimizer (from SGD to AdamW), text encoder (from LSTM to 2-layered LSTM+GRU), and embedding dimension of composed feature (from 512 to 2048) to maximize the result of a single model.}
\centering
\begin{tabular}{cc}
\hline
Training details & Validation \\
\hline \hline
Baseline (RTIC) & 33.24 \\
\hline
+ Replace optimizer & 34.89 \\
+ Increase embedding dimension & 36.79 \\
+ Replace text encoder & 37.72 \\
+ Spell correction & \textbf{38.22} \\
\hline
\end{tabular}
\label{tab:improvement_on_each_step}
\end{table}

\begin{figure*}[t!]
\centering
\includegraphics[width=1.0\textwidth]{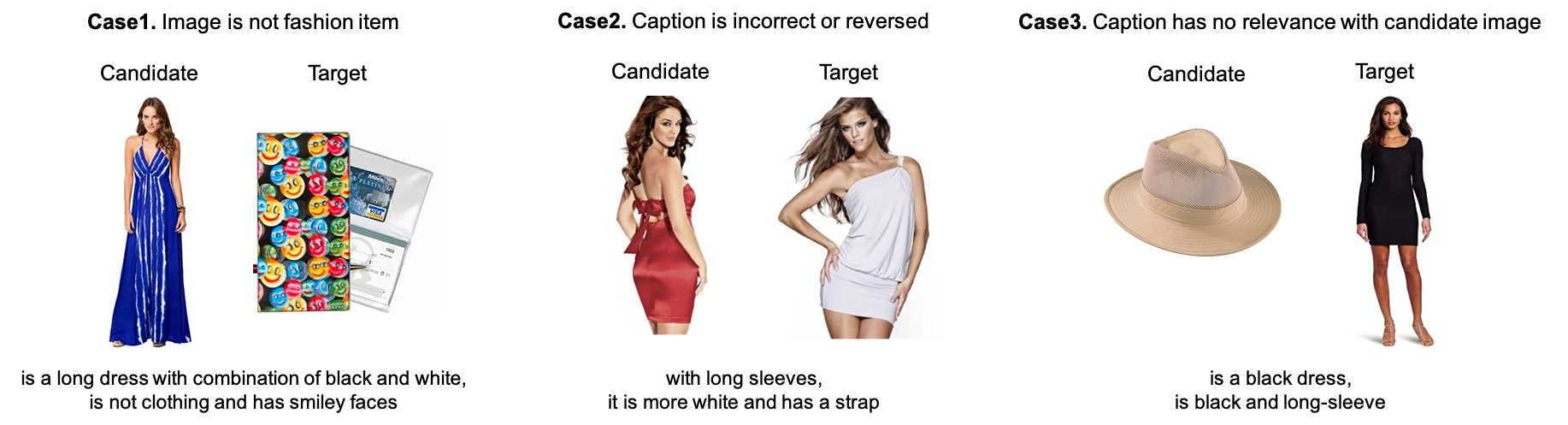}
\caption{Overview of Fashion-IQ dataset. The data are provided in pairs with a candidate image, target image, and caption. Although the caption should describe the relative differences between the target and the candidate image, some annotations were inappropriate as we inspected. For example, some images are not a fashion item (Case 1), the caption is incorrect or reversed (Case 2), or the target has no relevance with the candidate (Case 3).}
\label{fig:dataset_overview}
\end{figure*}

\section{Result}
The performance of our single model and the ensemble result on the validation set are shown in Table~\ref{tab:single_model_performance}. We trained RTIC with variants of text encoders. The SWEM~\cite{shen2018baseline} indicates that we encoded the sentence embedding as the concatenation of the average pool and the max pool of the word embedding by GloVe~\cite{pennington2014glove} instead of encoding a sentence by LSTM or GRU from scratch. Although the performance of the individual model does not exceed our best single model (\textit{RITC, 38.22}), the ensemble result is improved significantly (\textit{Ensemble 4 models, 45.05}). With the iterative process of the ensemble using models trained in various experimental settings, the score is maximized to 47.55. At last, we merge 80\% of validation pairs into train pairs and trained the additional models. The results of these models on the validation set are not compared in Table~\ref{tab:single_model_performance} since it could break the tendency between the test and the validation score. Finally, by adding the result of the additional models, the ensemble score on the test was marginally improved (46.16 $\rightarrow$ 48.02).

\section{Conclusion}
We described our approach for the Fashion-IQ Challenge 2020 and reported the result. To solve the task, we proposed a novel method, the residual text and image composer (RTIC), that learns the residual between the representations of the target and the candidate image conditioned on the text. Our best single model achieves 38.22 average recall on validation. Our final submission is an ensemble of multiple distinct models (TextOnly, Image retrieval TIRG, RTIC) trained on the combinations of various experimental settings as described in Figure~\ref{fig:main_diagram}. We introduced an iterative process for the ensemble that computes a weighted sum of the similarity score of single models and the existing best score, where the optimal weights are found using Bayesian Optimization. Our final submission is 48.02 on the test set, which is 25.6\% of a performance boost comparing to the best single model.

{\small
\bibliographystyle{ieee_fullname}
\bibliography{paper}
}

\newpage
\appendix

\section{Dataset overview}
We show the overview of the Fashion-IQ dataset and analyze the cases of wrong data pairs in Figure~\ref{fig:dataset_overview}.

\section{Things that did not work}
We have explored various tricks to improve a single model performance. However, some of them did not work as we expected.

\noindent
\textbf{10-Crop.}
As discussed in~\cite{lastyear1st}, we average the 10-crop features of a single model by cropping the given image into four corners and the central crop plus the flipped version of these. However, the performance was rather poor.

\noindent
\textbf{Object detection.}
The images were cropped using the fashion object detector finetuned on the DeepFashion2~\cite{ge2019deepfashion2} dataset. However, we observe that the performance is degraded in contrast to the result discussed in~\cite{lastyear3rd}. Our interpretation is that the images in Fashion-IQ dataset are low in resolution, which means that cropping object regions and resizing them into 224x224 would blur the input.

\noindent
\textbf{Use of more data.}
To increase the amount of train data, we created pairs of data from the FashionGen~\cite{rostamzadeh2018fashion} and the Fashion200K~\cite{han2017automatic} dataset and merged them into original Fashion-IQ pairs. Despite the train data increase, the single model performance was degraded tested on the Fashion-IQ validation set. However, we found that the ensemble result could be improved with model training on a different condition because of the model diversity.

\noindent
\textbf{Feature fusion.}
We tried diverse feature fusion methods\footnote{https://github.com/Cadene/block.bootstrap.pytorch} such as Tucker, MCB, MLB, BLOCK, and Mutan to examine if the feature fusion method works better than a simple concatenation. However, we could observe no evidence of improvement.

\noindent
\textbf{Bias to each domain}
After pretraining a model on the entire data pairs, we tried to finetune the model with a small $lr$ on each category (shirts, dress, and toptee). However, the result was degraded compared to the pretrained model. Furthermore, we observe that the single model result on each category does not always be identical even though it is trained with the same training parameters, which is caused by the randomness on data shuffling every epoch.

\end{document}